\documentclass[10pt, a4paper]{article}

\usepackage{lrec-coling2024} % this is the new style
\usepackage{scalerel}
\usepackage[usestackEOL]{stackengine}
\usepackage{multirow}
\usepackage{multicol}
\usepackage{listings}
\usepackage{lingmacros}
\usepackage{natbib}
\usepackage{multibib}
\usepackage{adjustbox}
\usepackage{comment}
\makeatletter
\def\@mb@citenamelist{cite,citep,citet,citealp,citealt,citepalias,citetalias}
\makeatother
\newcites{languageresource}{~}

\usepackage{graphicx}
\usepackage{wrapfig}
\usepackage{tabularx}
\usepackage{soul}

\usepackage{titlesec}
\titleformat{\section}{\normalfont\large\bfseries\center}{\thesection.}{1em}{}
\titleformat{\subsection}{\normalfont\bfseries\raggedright}{\thesubsection.}{1em}{}
\titleformat{\subsubsection}{\normalfont\normalsize\bfseries\raggedright}{\thesubsubsection.}{1em}{}
\renewcommand\thesection{\arabic{section}}
\renewcommand\thesubsection{\thesection.\arabic{subsection}}
\renewcommand\thesubsubsection{\thesubsection.\arabic{subsubsection}}
\usepackage{xcolor}
\usepackage{hyperref}
 \definecolor{darkblue}{rgb}{0, 0, 0.5}
  \hypersetup{colorlinks=true, citecolor=darkblue, linkcolor=darkblue, urlcolor=darkblue}

\usepackage{xstring}

\usepackage{booktabs}
\usepackage[inline]{enumitem}
\usepackage{arydshln}

\usepackage{color}

\usepackage{pifont}% http://ctan.org/pkg/pifont
\newcommand{\cmark}{\ding{51}}%
\newcommand{\xmark}{\ding{55}}%

\title{Common Ground Tracking in Multimodal Dialogue}
\name{Ibrahim Khebour\textsuperscript{1}, Kenneth Lai\textsuperscript{2}, Mariah Bradford\textsuperscript{1}, Yifan Zhu\textsuperscript{2}, Richard Brutti\textsuperscript{2},\\
 Christopher Tam\textsuperscript{2}, Jingxuan Tu\textsuperscript{2}, Benjamin Ibarra\textsuperscript{1}, Nathaniel Blanchard\textsuperscript{1}, Nikhil Krishnaswamy\textsuperscript{1}, and \\ James Pustejovsky\textsuperscript{2}}
 
\address{\textsuperscript{1}Colorado State University, Fort Collins, CO, USA\\
\textsuperscript{2}Brandeis University, Waltham, MA, USA\\
\{ibrahim.khebour, nkrishna\}@colostate.edu,jamesp@brandeis.edu}

\abstract{
Within Dialogue Modeling research in AI and NLP, considerable attention has been spent on ``dialogue state tracking'' (DST), which is the ability to update the representations of the speaker's needs at each turn in the dialogue by taking into account the past dialogue moves and history. Less studied but just as important to dialogue modeling, however, is ``common ground tracking'' (CGT), which identifies the \textit{shared} belief space held by all of the participants in a task-oriented dialogue: the task-relevant propositions all participants accept as true. In this paper we present a method for automatically identifying the current set of shared beliefs and ``questions under discussion'' (QUDs) of a group with a shared goal. We annotate a dataset of multimodal interactions in a shared physical space with speech transcriptions, prosodic features, gestures, actions, and facets of collaboration, and operationalize these features for use in a deep neural model to predict moves toward construction of common ground. Model outputs cascade into a set of formal closure rules derived from situated evidence and belief axioms and update operations. We empirically assess the contribution of each feature type toward successful construction of common ground relative to ground truth, establishing a benchmark in this novel, challenging task.
 \\ \newline  
}

\stackMath
\begin{document}

\maketitleabstract
 
\section{Introduction}
\vspace*{-2mm}
In the context of increasingly sophisticated interactions involving natural language dialogues with an AI, there is considerable attention being spent on ``Dialogue State Tracking'' (DST), which is the ability to update the representations of the speaker's (user's) needs at each turn in the dialogue, by taking into account the past dialogue moves and history. In this paper, we address the related but less-studied problem of {\it Common Grounding Tracking} (CGT), which identifies the shared belief space held by all of the participants in a task-oriented dialogue. We describe the procedure for training CGT models to both identify the current set of beliefs, as well as determine the level of evidence for each, to condition where the dialogue will go (the ``questions under discussion'', or QUDs). The goal is to provide a more informative snapshot of the dialogue situation, after each action in the task, to develop a policy incorporating shared beliefs in addition to past dialogue history.

A major challenge facing the development of computational models for multimodal interactions involves  tracking the intentions, goals, and attitudes of the participants \cite{cassell2000embodied,foster2007enhancing,kopp2010gesture,marshall2013theories,schaffer2019conversation,wahlster2006dialogue}. For task-oriented dialogues, just as important is the problem of identifying and tracking the {\it common ground} between participants \cite{clark_grounding_1991,traum1994computational,asher1998common,
dillenbourg2006sharing}.

In this work, we specifically: (a) identify both communicative expressions (speech, gesture) and jointly perceived actions in a multi-party dialogue, in order to convert them into propositional content; and (b) add them to a dynamic data structure we call the {\it Common Ground Structure (CGS)}. This consists of three parts: {\bf FBank}, a set of facts that are  assumed to be known by the group; an {\bf EBank}, a set of evidences available to the group;  and {\bf QBank}, the ``questions under discussion", a set of topics remaining to be discussed in order to solve the task.  

In total, this work encompasses three novel contributions:

%\todo{Redundant with Section 4.  Remove either here or there.}
%The Weights Task is a collaborative problem-solving task in which participants work together to deduce the weights of blocks by making the correct block comparisons on a scale. This task involves multiple modalities, of which we are currently annotating speech, gesture, and action. We aim to track the group's collective acceptance by supplying another layer of annotation on top of our existing ones, augmenting them with what we call "common ground annotations (CGA)". 

\vspace*{-2mm}
\begin{itemize}
    \item A challenging new task: multimodal common ground tracking, with a formal model of common ground in a shared, situated task;
    \vspace*{-2mm}
    \item A novel incorporation of the formal model into an automated pipeline that tracks the evolution of group common ground over time;
    \vspace*{-2mm}
    \item An augmentation of the Weights Task Dataset \cite{khebour_ibrahim_2023_8384960} with gesture, action, and common ground annotations, to enable the operationalization of our formal model.
\end{itemize}
\vspace*{-2mm}

Our code may be accessed at \hyperlink{https://github.com/csu-signal/Common-Ground-detection}{https://github.com/csu-signal/Common-Ground-detection}

\begin{figure}
    \centering
    \includegraphics[width=.48\textwidth]{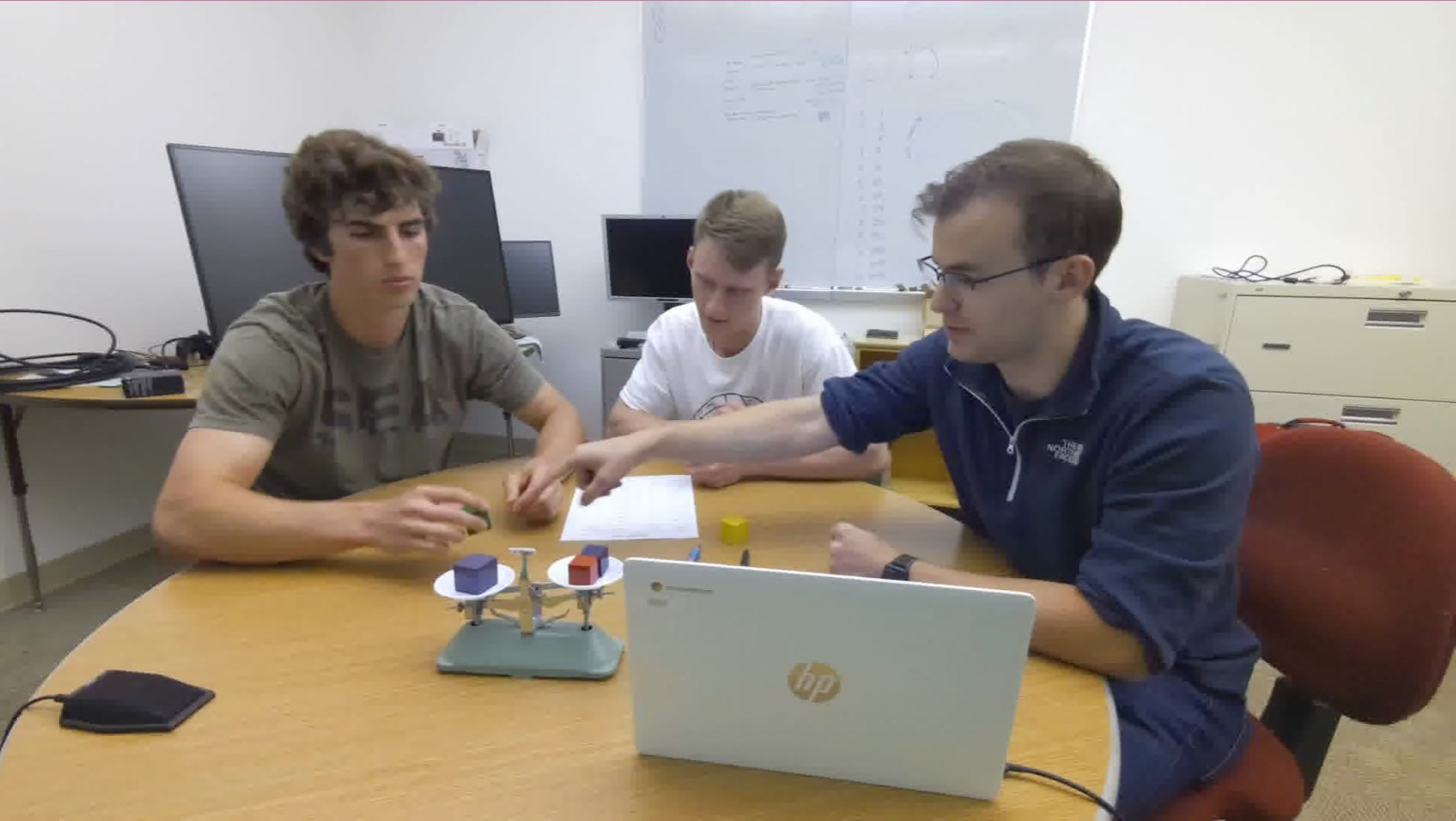}
\vspace*{-2mm}
    \caption{\label{fig:wtd-sample}Sample still from the Weights Task Dataset showing communication with multiple modalities. The accompanying utterance at this time is ``Put the twenty on there; take off a ten''.}
\vspace*{-2mm}
\end{figure}

\vspace*{-2mm}
\section{Related Work}
\vspace*{-2mm}

The present work draws on several diverse areas of research, from modeling common ground in HCI and HHI, and Dialogue State Tracking, to the role of gesture in multimodal interactions. 

When engaged in  dialogue, our shared understanding of both utterance meaning (content) and the speaker's meaning in a specific context (intent), involves the ability to link these two in the act of situationally grounding meaning to the local context, what is  typically referred to as ``establishing the common ground" between speakers \cite{grice1975logic,clark_grounding_1991,stalnaker2002common,asher1998common,traum2003information}.  The concept of common ground refers to the  set of shared beliefs among participants in a Human-Human interaction (HHI) ~\cite{markowskaformal,traum1994computational,hadley_review_2022}, as well as HCI \cite{krishnaswamy2019generating,ohmer2022emergence} and HRI interactions \cite{kruijff2010situated,fischer2011people,scheutz2011toward}. ~\citet{del2022rewriting}  have recently employed the notion of common ground operationally to identify and select relevant  information for conversational QA system design. 
~\citet{stewart_multimodal_2021} and \citet{bradford2023automatic} both study human-human collaboration through the lens of an AI agent. 

%Components of an AI system for use in collaborative environments include, but are not limited to, speaker diarization~\cite{park2022review}, modeling dialogue acts~\cite{enayet2022analysis, mezza2022multi, trippas2020towards}, and situated grounding~\cite{krishnaswamy2019situated, ruckert-etal-2022-unified}. Modeling {\it common ground} between users and agents is critical to dialogue systems~\cite{furuya-etal-2022-dialogue}, but common ground is also a necessary component of multiparty human-human interactions, and automatically modeling common ground between multiple humans remains a challenge. Building common ground requires both presentation and acceptance of information, which can manifest in varihttps://www.overleaf.com/project/64ee23f70c4d4e941d47d6c7ous ways~\cite{clark_grounding_1991, fusaroli2017measures}.

Dialogue state tracking (DST) aims to estimate the current dialogue state or belief state of the users during the conversation \cite{budzianowski-etal-2018-multiwoz,liao2021dialogue,jacqmin2022you}. Current
 DST models can be categorized into three types: fixed ontology~\cite{henderson-etal-2014-word,mrksic-etal-2017-neural,Chen2020SchemaGuidedMD}, open vocabulary~\cite{gao-etal-2019-dialog,hosseiniasl2022simple,wu-etal-2019-transferable} and hybrid methods ~\cite{goel19_interspeech, zhang-etal-2020-find,heck-etal-2020-trippy}. 
Recently, pretrained language models have been widely used to model slot relations, while Graph attention networks (GATs) have been used to model the hierarchical structure of DST,  enabling the incorporation of semantic compositionality, cross-domain knowledge sharing and coreference~\cite{lin-etal-2021-knowledge, li-etal-2021-generation,cheng-etal-2020-conversational}. 

%To get a comprehensive understanding of dialogue, many researchers transform DST into other tasks.~\citet{gao2020machine} uses reading comprehension task (RC) techniques for state tracking.~\citet{dai-etal-2021-preview} proposes to use curriculum learning (CL) to better leverage both the curriculum structure and schema structure for task-oriented dialogues to capture the rich structural information in a dataset. To get a comprehensive understanding of dialogue, many researchers transform DST into other tasks.~\citet{gao2020machine} uses reading comprehension task (RC) techniques for state tracking.~\citet{dai-etal-2021-preview} proposes to use curriculum learning (CL) to better leverage both the curriculum structure and schema structure for task-oriented dialogues to capture the rich structural information in a dataset.

Understanding the role of nonverbal behavior in multimodal communication has long been a research interest in HCI, but has recently taken on new interest within CL and the broader AI community.  Gestures offer an array of unique dimensions in communication, ranging from denoting situational references to indicating specific spatial locations or even conveying manner and  orientation ~\cite{inproceedings,book,article,doi:10.1146/annurev.anthro.26.1.109,article2, book2}. 
Gesture AMR (GAMR) ~\cite{brutti-etal-2022-abstract} considers gestures that convey the same propositional content and intentionality as speech acts. Gesture may have meaning on its own, or it may enhance the meaning provided by the verbal modality~\cite{book3,krishnaswamy2020formal}. 
Also critical to multimodal dialogue is human action, which in addition to communicating deictic and bridging information can also make lasting changes to the world, affecting the common ground ~\cite{tamannotating}. Much work has been done to facilitate action identification from video ~\cite{sigurdsson2016hollywood} ~\cite{gu2018ava} ~\cite{li2020avakinetics} as well as to annotate specific semantic roles ~\cite{sadhu2021visual}.

\citet{di2021cutting} implement dynamic belief sets as graphs, which we do not do explicitly. However, such an approach is theoretically and computationally compatible with ours, as the result (post-condition) of a public announcement or observed action can act as the preconditions for promoting QUDs to evidence, or evidenced propositions to strong beliefs, leading to a natural interpretation of common ground tracking as a graph.
%\todo{Address this paper: "Cutting melted butter? Common Ground inconsistencies management in dialogue systems using graph databases." IJCoL. Italian Journal of Computational Linguistics 7.7-1, 2 (2021): 157-190.}

%\nk{Rebuttal:
%Thank you for sharing this very interesting paper, which is very compatible with the approach presented in our paper both theoretically and computationally. While we do not implement our dynamic belief sets explicitly as graphs, they are certainly interpretable as such. In fact, given that the result (post-condition) of a public announcement or observed action can act as the preconditions for promoting QUDs to evidence, or evidenced propositions to strong beliefs, there is a natural interpretation of our approach as a common ground graph.}

\vspace*{-2mm}
\section{Dataset}
\label{sec:data}
\vspace*{-2mm}

The Weights Task \cite{khebour_ibrahim_2023_8384960} is a collaborative problem-solving task in which groups of three work together to deduce the weights of differently-colored blocks by making comparisons of block weights using a balance scale. In this activity, the group has a balance scale and five blocks of various colors, sizes, and weights. They are told the weight of one block and must identify the weights of the remaining blocks and, eventually, the algebraic relation between them, which is an instance of the Fibonacci Sequence~\cite{sigler2002fibonacci,bonacci1202liber}. Due to the co-situated nature of the task and its inclusion of physical objects and reasoning about their properties, this task involves communication in multiple modalities, such as language, gesture, and action (see Fig.~\ref{fig:wtd-sample}), meaning that knowledge is shared using multiple communicative channels. The Weights Task Dataset (WTD) comes with automatic and human transcriptions of the speech, as well as gesture annotated using Gesture-AMR (GAMR)~\cite{brutti-etal-2022-abstract}, and collaborative problem solving (CPS) indicators according to the framework of \citet{sun2020towards}. All groups successfully deduce the correct block weights, giving a consistent end state against which to assess our models.  

\subsection{Example Dialogue}
\label{ssec:example}

\begin{figure}[h!]
    \centering
    \includegraphics[width=.48\textwidth]{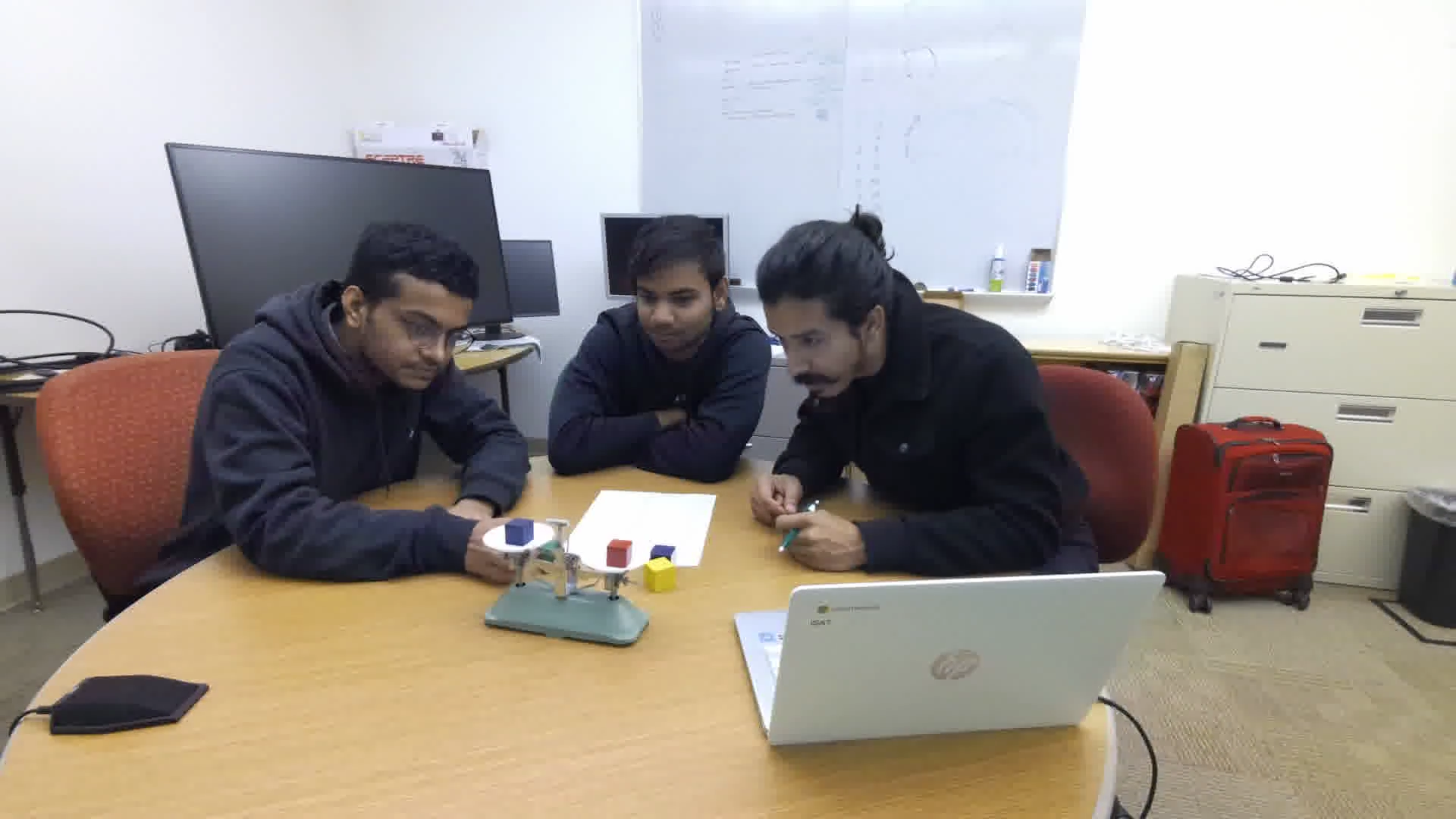}
\vspace*{-2mm}
    \caption{\label{fig:example}Example dialogue. Participant 3 (right) says ``looks like they're fairly equal" after placing the red and blue blocks on different sides of the scale. We refer back to this example elsewhere in the paper.}
\vspace*{-2mm}
\end{figure}

In the WTD, participants are canonically indexed from 1--3, left to right. In Fig.~\ref{fig:example}, Participant 3 makes a statement that the red and blue blocks are ``fairly equal'', which is interepreted as an assertion of belief that $red = blue$. Participant 1 give a qualified assent to this through the utterance ``yeah, I suppose,'' meaning that at this point, $red = blue$ and other necessarily entailed propositions can be considered part of the common ground (for example, if we had established that $red < yellow$, then $blue < yellow$ also becomes part of the common ground).

\vspace*{-2mm}
\section{Common Ground in Dialogue}
\label{sec:CG}
\vspace*{-2mm}

Here we assume the context of a multi-participant, task-oriented conversation, involving communication by multiple content-generating modalities (language, gesture) and mutually interpretable non-verbal behaviors (e.g., actions) \cite{kruijff2010situated,pustejovsky2021embodied}. 
To this end, we need a data structure representing the common ground in such a context, that can be dynamically updated throughout the dialogue. We adopt a version of a Dialogue Game Board (DGB), as developed in \citet{ginzburg2012interactive}. 

%In order to model  how participants in a multimodal dialogue update their beliefs, we will adopt an evidence-based model of an agent's epistemic stance towards a proposition \cite{van2011dynamic}. 
 
Because of the evolving and dynamic nature of co-interactive dialogue and situated actions, following \citet{van2014evidence} and \citet{pacuit2017neighborhood}, we adopt an {\it evidence-based model} of belief, where our commitments to propositions describing situations or facts are not binary, but are graded, where they can weaken or strengthen depending on available {\it evidence} for them as the dialogue progresses.

First, however, we define the minimal structure of a task-oriented interaction as a sequence,  $D$,  of dialogue steps, where each move in the dialogue takes it into another situation or state. 
%When considering multiple modalities of communication, along with the modality of action itself, we can generalize $D$ to a multimodal  dialogue ($D_{m}$). We will define the transitioning step from one situation to the next, as a generalization of a dialogue step. 
Let $P = \{p_1,p_2,p_3\}$, be the participants in  our dialogue. From any situation $s_k$, we define a $D$ move, $m_i$, as $m_i = (p_j, C_j, s_{k+1})$: participant $p_j$ performs a communicative act $C_j$, bringing the multimodal dialogue into situation $s_{k+1}$. The $D$ can be defined as the sequence of these moves:
$D$ = $m_1,\ldots,m_n$.

 Here our interest is in tracking the situation content resulting from each move:   the set of propositions that captures the current state of the world, the current progress towards a goal, or the status of a task. In addition, it captures    the current questions under discussion and beliefs  in the dialogue.   

Given these considerations, we identify three components  for tracking common ground in dialogue: a minimal static model of degrees of belief; a data structure distinguishing the elements of the agents' common ground that are being tracked; and a dynamic procedure which updates this structure, when new information and evidence is available to the agents. We consider each of these in 
turn below.

\vspace*{-2mm}
\subsection{Evidence-based Belief}
\vspace*{-2mm}

\citet{pacuit2017neighborhood} provides a model for evidence-based
belief, where agents  obtain evidence in favor of a proposition, $\varphi$, and can , to eventually believe $\varphi$.
We adopt a simplified model of the evidence-based Dynamic Epistemic
Logic (EB-DEL) as developed in
\citet{van2014evidence} and \citet{pacuit2017neighborhood}. We define a model as a tuple,
$\mathcal{M} = (W,E,V)$, where

\enumsentence{
a. $W$ is a non-empty {\it set of worlds}; 
\\
b. $E\subseteq W \times \wp(W)$ is an {\it evidence relation};
\\
c.   $V$$: At \rightarrow \wp(W)$, is a  {\it valuation function}.
}

\noindent Let  $E(w)$ denote  the set $\{ X \;|\; wEX \}$, the worlds accessible to $w$ through the evidencing relation, $E$. The evidence-based epistemic language, $\mathcal{L}$, will be the set of formulas generated by the grammar below:

\enumsentence{
$p \; | \; \neg \varphi \; | \;\varphi \wedge \psi \; | \; [E]\varphi \;| \; [B]\varphi \; | \; [A]\varphi $

}

\noindent  We  distinguish the situation where an agent has ``evidence in favor of" a proposition $\varphi$, as $[E]\varphi$. Because an agent can have evidence for propositions that convey contradictory information, she can consider both $[E]\varphi$ and $[E]\neg \varphi$. 
This corresponds to an agent having multiple neighborhoods, $X$, that are each evidenced in their unique way by $w$. However, consider the set of non-contradictory worlds as a unique subset of $X$, one which has what \citet{van2011dynamic} refer to as the {\it finite intersection property (fip)}. This property allows us to identify a neighborhood of accessible worlds with non-contradictory propositional content. When this occurs, we say  an agent has {\it belief} in a proposition, $[B]\varphi$.  Finally, the universal modality  is considered ``knowledge" of a proposition, $[A]\varphi$. 

\vspace*{-2mm}
\subsection{Common Ground Structure}
\vspace*{-2mm}

%Consider Now consider how information is updated through dialogue and observation of actions, that is, the dynamics of what will be learned through the multimodal interactions. 

Capturing situational state information in a task-oriented dialogue is critical for reflecting current common ground as well as predicting future dialogue moves \cite{traum2003information,schlangen2011general,zhang2020recent,jacqmin2022you}. For our present purpose,  we adopt the notion of a {\it Dialogue Game Board} \cite{ginzburg1996dynamics,ginzburg2012interactive}, modified to reflect the varying degrees of evidence associated with propositions under discussion.  
A Common Ground Structure, $cgs$, is a triple, $(QB,EB,FB)$, consisting of:

\enumsentence{
a. Questions Under Discussion ({\sc QBank}): set of topics or unknowns that need to be answered to solve the task; \\
b.  Evidence ({\sc EBank}): set of propositions for which there is some evidence they are true; 
\\
c. Facts ({\sc FBank}): set of propositions believed as true by all participants.
}
 
%The QBank is initialized with all the possible questions of interest that can be posed regarding any object in the domain. However, for our common ground tracking task, we limit the QUDs to only those concerning the task, i.e., the blocks and the scale. QUDs involving participants are not included in this model. 

 The task begins with a set of unknowns referred to as the ``Questions under Discussion'' (QUDs). For this implementation, we create a finite model, including a finite model of questions. For all objects in the domain relating to the task, questions are generated for each relation implicated in the task for that object. For example, in the Weights Task, the goal is to identify the weights of five distinct blocks, and then the algebraic relation between them, i.e., the Fibonacci sequence. 
The weight of a block ranges between 10 and 50 grams, in 10 gram intervals. Hence, for each block in $B$, where $B$ = $\{red,blue,yellow,green,purple\}$, we have five possible values, expressed as yes/no questions. Hence, initialization of the QBank results in the following set:
 
\enumsentence{
QBank = \\
 $\{Eq(r,10)?,\ldots , Eq(r,50)?,
  \ldots  , Eq(p,10)?,$ \\$ \ldots Eq(p,50)?\}$

}

\noindent
At the outset of our dialogue, we set both EBank and FBank to nil, since no task-relevant propositions have been established as commonly evidenced or believed.  In the next section, we address the task of determining how information is updated in the dialogue, thereby changing the common ground.

\vspace*{-2mm}
\subsection{Updating the Common Ground}
\label{ssec:updating-cg}
\vspace*{-2mm}

%{\bf Introduce the update operators}

Given the epistemic logic presented above, we introduce the mechanisms that update the information state within a dialogue. Following \citet{plaza1989logics} and subsequent developments of Public Announcement Logic \cite{baltag2016logic}, we introduce  a new  operator to the model, the announcement operator, $!$. 
Public announcements are statements that are made to all agents, and
after the announcement, all agents know that the statement has been
announced and that it is true.

If 
$[!\varphi ]$ represents the act of announcing  
$\varphi$,  then 
$[!\varphi ]\psi$
 means ``after  $\varphi$ is announced, then $\psi$ is believed to
 be the case."

In order to distinguish evidence for $\varphi$ from belief in $\varphi$, we relativize the impact of a statement to the context within which it is uttered. Let us interpret $[!\varphi ]\psi$ as follows. 

\eenumsentence{
\item {\it Update with Evidence}:\\
$[!\varphi ][E]\psi$: 
Given the announcement of $\varphi$, there is evidence for $\psi$; 
\item {\it Update with Belief}: \\ $[E]\varphi \rightarrow [!\varphi ][B]\psi$: Belief in $\varphi$ is conditionalized on $\varphi$'s announcement in the prior context of evidence for $\varphi$. 
}\label{updates}

Semantically, an
update  represents the state of affairs after an announcement. This
entails 
transforming the current  model by removing all states where the announced formula
is false. With evidence distinguished from belief/knowledge, we also update the evidence function, where $[!\varphi ]$:
\enumsentence{
a. Updates the worlds: $W' = W \cap \varphi$
\\
b. Updates the Evidence function: $E'(w) = E(w) \cap \varphi$ \\
c.  $(M,w) \; \models \; \varphi$ implies $(M|_{\varphi},w) \; \models  \; [E]\psi$
}

\noindent 
This update actually changes the underlying evidence sets themselves. The announcement is taken as a piece of direct evidence. Hence, to capture that the announcement of $\varphi$
becomes evidence and not just belief, the evidence sets for each agent get restricted (or updated) to reflect the worlds where  $\varphi$ is true. Subsequently, the belief function will then naturally adjust based on the new evidence sets.

%\noindent We say that: 
%$(M,w) \models  \varphi$ implies %$(M|_{\varphi},w)  \models [E]\psi$.

 Operationally, after (\ref{updates}a) is run, the  model is relativized to evidencing neighborhoods, where $\varphi$ is true. This corresponds to moving a proposition from QBank to the EBank. 
Then, if the same proposition is ``announced" again, as with an {\it ACCEPT} move, then (\ref{updates}b) promotes that proposition from the EBank to the FBank. 

In Section \ref{ssec:closure}, we illustrate how these updates are applied when running over the output of the move classifier, in order to determine the content of the current common ground.

\vspace*{-2mm}
\section{Annotation}
\label{sec:anno}
\vspace*{-2mm}

We augmented the existing WTD annotations with dual annotation of GAMR, and participant actions using VoxML~\cite{pustejovsky2016LREC}. Finally, we also tracked the group's collective surfacing of evidence and acceptance of task-relevant facts by supplying another layer of ``common ground annotations'' (CGA):

Annotation in the dialogue involves identifying categories relating to the cognitive state of participants to actions and knowledge, concerning the task. This includes the following categories:
 (a) \textit{OBSERVATION}: participant $P_i$ has perceived an action, $a$; (b)
 \textit{INFERENCE}: deduction from $\varphi$;  (c)  \textit{STATEMENT}: announcement of evidence
$\varphi$;  (d) \textit{QUESTION}: introducing role interrogative relating to $\varphi$; 
    (e)  \textit{ANSWER}; supplying filler to question about $\varphi$; 
   (f)   \textit{ACCEPT}: agree with evidence $\varphi$;
 (g)   \textit{DOUBT}: disagree with evidence for $\varphi$. 
   %  Recommendation; an action should be performed. 

In the example from Sec.~\ref{ssec:example}, Participant 3's utterance would be considered a \textit{STATEMENT} of the proposition $red=blue$ while Participant 1's utterance would be an \textit{ACCEPT} of that proposition. Participant 3 subsequently says ``that's 20, these two [referring to the red and blue blocks] are 10'' (\textit{STATEMENT} of proposition $red=10 \wedge blue=10$), to which Participant 1 says ``wait, let's see'' signaling a \textit{DOUBT} in $red=10 \wedge blue=10$. Therefore at this stage of the dialogue, $red=blue$ can be considered an agreed-upon fact (element of FBank), but none of the participants have accepted that $red=10 \wedge blue=10$, so that proposition is still only an element of EBank. This example illustrates the subtleties captured through the annotation.

GAMR, action, and common ground annotations were all dually-annotated. GAMR annotations achieved a SMATCH-F1 score of 0.75.  Action annotation achieved an F1 score of 0.67 and Cohen's $\kappa$ of 0.59. CGA achieved F1 of 0.54 and Cohen's $\kappa$ of 0.50.  Each was adjudicated by an expert to produce the gold standard.

\vspace*{-2mm}
\section{Experiments: Modeling Common Ground Tracking}
\label{sec:exp}
\vspace*{-2mm}

Our experimental pipeline consists of 3 primary components: a {\it move classifier}, which predicts which cognitive state is being expressed in an utterance (Sec.~\ref{sec:anno}); a {\it propositional extractor}, which may either consult a dictionary of expressed propositions that is collected from the annotated data with all modalities considered, or may automatically extract the propositional content of an utterance through vector-similarity methods; and a set of {\it closure rules} that unify the cognitive state and propositional content and update the status of QBank, EBank, and FBank.

Our primary metric for the entire pipeline is S\o rensen-Dice coefficient (DSC) \cite{sorensen1948method,dice1945measures}. DSC indicates how much the set of propositions extracted by the model matches the set of propositions in the ground truth. It also normalizes for the size of the samples being compared, as the cardinality of the different banks may fluctuate widely as the task proceeds.  DSC can also be evaluated as a group proceeds through the task, or averaged over a single group.  Since the groups have different numbers of utterances, and hence moves, when aggregating across groups to calculate DSC over time, we pad the length of the shorter groups out to the maximum length by copying the final state of the banks, assuming a ``steady state'' in the common ground once the task is finished.

\vspace*{-2mm}
\subsection{Preprocessing}
\label{ssec:prep}
\vspace*{-2mm}

We first mapped the annotated data to the ``oracle'' (manually-segmented) utterances in the WTD~\cite{terpstra2023good}. If more than one annotation for a given modality was present in the same utterance, we used the one that had the biggest overlap with the oracle utterance.

We encoded the manually transcribed utterances in the WTD into embedding vectors using BERT ({\tt base-uncased})~\cite{devlin-etal-2019-bert}, and extracted the 768D {\tt [CLS]} token embedding from the final encoder layer. Following \citet{bradford2023automatic}, we processed the audio into 88D prosodic features using openSMILE~\cite{eyben2010opensmile}. The CPS indicators for an utterance were transformed into their corresponding high-level {\it facets} according to the \citet{sun2020towards} framework, and encoded as 3D one-hot vectors.

{\bf GAMR representations} were featurized as $k$-hot encodings of size 81. The first 4 components describe the gesture type (icon, gesture-unit, deixis and emblem) along with a fifth component that represents {\it and} in cases where 2 types are annotated. Following this is one hot encoding of the GAMR {\tt ARG0} (gesturer), then a $k$-hot encoding vector of components of the GAMR {\tt ARG1} (gesture content, such as object of deixis). Given the vocabulary of items in the data, this comprises 68 dimensions. Finally, GAMR {\tt ARG2} (gesture recipient) was represented by a one-hot encoding of size 5 (group/researcher/1 of 3 participants). As more than one participant can have a GAMR annotated for them over the same utterance, we allow for 1 GAMR feature vector per participant, resulting in a total GAMR feature size of 243.

{\bf Action annotations} comprise scale actions, which were vectorized as a one-hot representation of the scale status (left/balanced/right), and participant actions. Participant actions comprise a 2D representation of ``lift'' vs. ``put'', a one hot encoding for the block being acted upon, a 2D one hot encoding of ``in'' and/or ``on'', and a 2nd one-hot object representation of block, scale, or table (the destination of the action, where applicable). A participant's action vector is of size 25 ($\times$ 3 for 3 participants), resulting in a total action feature size of 78, including the scale actions.

To facilitate propositional extraction, we decontextualized each utterance from the dialogue context, inspired by the {\it dense paraphrasing} method \cite{tu-etal-2022-competence,dp-iwcs} that rewrites a textual expression to reduce ambiguity and make explict the underlying semantics. We filtered the utterances containing at least one pronoun from a predefined set, and had annotators identify the blocks denoted by the pronouns, if any, based on the aligned actions and video frames. Utterances were dually annotated (Cohen's $\kappa$ = 0.88) and adjudicated by an expert. Utterances were paraphrased by replacing the pronouns with the adjudicated annotation of the block colors (e.g., \textit{\textbf{they [red block and blue block]} are probably equal}). 

The annotated dataset presents a number of challenges related to sparsity, imbalance, and cross-group diversity.  The individual feature channels either capture a single communicative modality or cross-cut two or more (viz. prosodic and CPS features). For input to the model, we concatenate the features of each utterance to the $w$ previous utterances. We remove utterances with no CGA, unless they fall within the $w$-utterance context window of an utterance with CGA. Of the 1,822 utterances in the dataset, only 271 have any common ground annotation, and these annotations are heavily biased toward the {\it STATEMENT} class (195, vs. 61 {\it ACCEPT}s and 15 {\it DOUBT}s).

\vspace*{-2mm}
\subsection{Move Classifier}
\label{ssec:move}
\vspace*{-2mm}

\begin{figure}[h!]
    \centering
    \includegraphics[width=.48\textwidth]{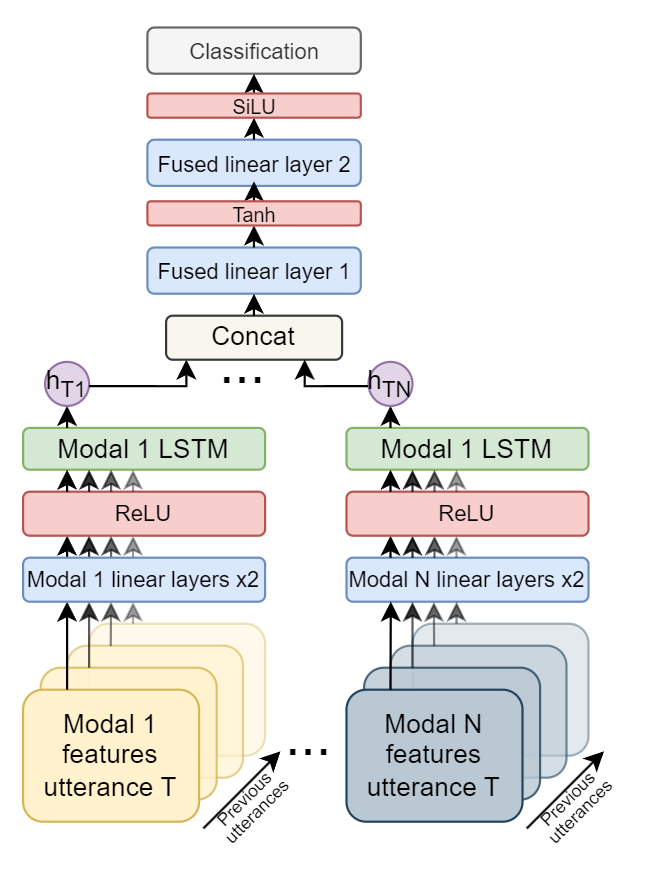}
\vspace*{-2mm}
    \caption{Move classifier architecture.}
    \label{fig:model-arch}
\vspace*{-2mm}
\end{figure}

The move classifier is a multimodal LSTM-based model, intended to capture contextual information that conditions the sequence of cognitive states in a dialogue. Each utterance, including a prior context of $w=3$ previous utterances, was processed through two linear layers (256 and 512 units) followed by ReLU activation and an LSTM block of 512 units.  The final hidden states of the LSTM block for each modality of interest were concatenated and passed through a 512-unit linear layer, $tanh$, another 512-unit linear layer, and SiLU before the classification layer. Fig.~\ref{fig:model-arch} shows this architecture.

We optimized for the detection of {\it STATEMENT}, {\it ACCEPT}, and {\it DOUBT}.  To alleviate imbalance during training, we augmented the data with SMOTE~\cite{chawla2002smote}.  We trained using Kaiming initilization with a uniform distribution~\cite{he2015delving}. All layers except the classification layer are trained using a triplet loss with a margin of 1 \cite{balntas2016learning} for 200 epochs and a learning rate of $10^{-4}$. Subsequently the entire model was trained using cross-entropy loss and a learning rate of $10^{-3}$ for 100 epochs, and for 200 further epochs with a learning rate of $10^{-4}$.  Hyperparameters were fixed using a search with one group held out as validation and one group as test, after which each group was held out in turn while an instance of the model is trained on the other 9 groups, for evaluation on the held-out test group.  We ended up with ten trained instances of the archicture, one for each group.

\vspace*{-2mm}
\subsection{Propositional Extractor}
\label{ssec:prop}
\vspace*{-2mm}

In additional to the cognitive state expressed by the utterance, we also needed to retrieve the task-relevant propositional content expressed relative to the QUDs. We used two methods for this:
\begin{enumerate*}[label=\arabic*)]
    \item {\bf CGA} (\textit{Common Ground Annotation}): We automatically mapped the statement IDs to the propositions expressed as captured in the common ground annotation. Upon move prediction, we consulted this mapping to retrieve the propositional content to be associated with the move in the common ground update. Because annotators had access to the video channel and all other modalities when annotating the propositions expressed, this method is a multimodally-informed method of propositional extraction.
    \item {\bf DP} (\textit{Dense Paraphrase}): We encoded the dense paraphrase of the input utterance through BERT ({\tt base-uncased}). Stop words were filtered out before encoding. The stop words came from a standard list, augmented with words that occured in the transcriptions in 5 or fewer bigrams \textit{and} are not number words, color words, or words describing equality or inequality. 
    BERT vectors were computed by summing over the last 4 encoder layers and taking the average of the {\tt [CLS]} token vector and all individual token vectors in the utterance. Upon move prediction, we chose the proposition whose similarly-encoded BERT vector had the highest cosine similarity to the utterance embedding. Only text and a language model were used in this method, making it unimodal. However, it is important to note that the annotators of the dense paraphrased utterance still had access to the video channel when determining which objects were the denotata of demonstrative pronouns, meaning that is some distant signal from the multimodal data reflected here.
\end{enumerate*}

These two methods provide an additional axis of comparison when computing the common ground and allow us to measure the performance gain provided a targeted annotation that directly takes into account all modalities vs. a method using only text and a language model.

\vspace*{-2mm}
\subsection{Closure Rules}
\label{ssec:closure}
\vspace*{-2mm}

In order to determine the contents of the CG banks after each utterance, we developed a set of closure rules consistent with the epistemic model presented in Section \ref{sec:CG}. These rules describe how utterances (specifically, \textit{STATEMENT}s and \textit{ACCEPT}s) affect what is known about the weights of the blocks, and whether certain possibilities are more or less likely than others. When the task begins, the set of possibilities for each block is initialized to $\{10,20,30,40,50\}$, with no evidence for or against any of those possibilities (i.e., the {\it evidence\_for} and {\it evidence\_against} sets are empty).

Given a \textit{STATEMENT} or \textit{ACCEPT}, we first parsed the propositional content of the utterance into one or more atomic propositions, $p \in At$, where an atomic proposition consists of a block name, a relation ($=$, $<$, $>$, or $\neq$), and a right-hand side. The right-hand side can either be a weight $\in\{10,20,30,40,50\}$, a block name, or a set of block names connected by $+$. Atomic propositions generally update knowledge about the block on the left-hand side of the relation; the exception is if there is a single block on the right-hand side, and less is known about the right-hand-side block (as measured by the relative sizes of the possibility and evidence sets), in which case that block's possibilities are updated instead.

Then for each atomic proposition, we updated the knowledge associated with the relevant block, according to the move type:

{\bf STATEMENT}s add evidence for or against certain weights. Statements of propositions ``block $=$ weight" directly add that weight to the {\it evidence\_for} set; e.g., $[!Eq(b,10) ][E]Eq(b,10)$.
For other statements, we compute the set of weights inconsistent with that statement, and add them to the {\it evidence\_against} set.

{\bf ACCEPT}s remove weights (specifically, those inconsistent with the proposition) from the set of possibilities (and both evidence sets).

Then, from the contents of the possibility and evidence sets for each block, we generated the contents of the CG banks:

If there was only one possibility for the weight of a block, ``block $=$ weight" was added to \textbf{FBank}; e.g., $[E]Eq(b,10) \rightarrow [!Eq(b,10) ][B]Eq(b,10)$.

Otherwise, for those weights in the block's {\it evidence\_for} set, we added ``block $=$ weight" to \textbf{EBank}. Similarly, for those weights in the block's {\it evidence\_against} set, we added ``block $\neq$ weight" to EBank. Inequalities for which evidence existed also were added to EBank.

For the remaining weights (not in either evidence set) in the set of possibilities for the block, we added ``block $=$ weight?" to \textbf{QBank}.

Considering the example in Sec.~\ref{ssec:example}, since the group already knows that $red = 10$ at that point, once $red=blue$ is accepted as a fact, the closure rules also elevate $blue=10$ to the same epistemic status as $red=10$. All other possibilities for blue block's weight are also removed from both evidence sets.

The ground truth contents of the CG banks were computed by running the closure rules directly over the annotated data.

\vspace*{-2mm}
\section{Results}
\label{sec:results}
\vspace*{-2mm}

\begin{table*}[h!]
    \resizebox{\textwidth}{!}{
    \begin{tabular}{lllllllllll}
\toprule
 & {\bf Group 1} & {\bf Group 2} & {\bf Group 3} & {\bf Group 4} & {\bf Group 5} & {\bf Group 6} & {\bf Group 7} & {\bf Group 8} & {\bf Group 9} & {\bf Group 10} \\
\midrule
 \multicolumn{11}{c}{All modalities} \\
\midrule
{\bf QBank} & 0.777 & 0.663 & 0.811 & 0.841 & 0.575 & 0.868 & 0.845 & 0.834 & 0.987 & 0.551 \\
{\bf EBank} & 0.250 & 0.574 & 0.709 & 0.926 & 0.391 & 0.734 & 0.793 & 0.063 & 0.985 & 0.250 \\
{\bf FBank} & 0.425 & 0.480 & 0.418 & 0.348 & 0.318 & 0.315 & 0.637 & 0.574 & 0.000 & 0.794 \\
{\bf F $\cup$ E} & 1.000 & 0.864 & 0.939 & 0.866 & 0.875 & 0.880 & 1.000 & 0.600 & 0.996 & 0.903 \\
\midrule
 \multicolumn{11}{c}{Language only} \\
\midrule
{\bf QBank} & 0.767 & 0.911 & 0.829 & 0.817 & 0.514 & 0.868 & 0.972 & 0.834 & 0.987 & 0.392 \\
{\bf EBank} & 0.344 & 0.713 & 0.712 & 0.812 & 0.335 & 0.691 & 0.904 & 0.049 & 0.985 & 0.262 \\
{\bf FBank} & 0.000 & 0.528 & 0.501 & 0.045 & 0.165 & 0.372 & 0.825 & 0.526 & 0.000 & 0.000 \\
{\bf F $\cup$ E} & 1.000 & 0.922 & 0.925 & 0.832 & 0.959 & 0.799 & 0.967 & 0.585 & 0.996 & 0.827 \\
\bottomrule
	\end{tabular}}
\vspace*{-2mm}
	\caption{\label{tab:avg_sdc-all_vs_bert-cga}Average DSC per group over all CG banks, comparing multimodal features and language only features.  Propositions are extracted using the CGA method.}
\vspace*{-2mm}
\end{table*}

\begin{table*}[h!]
    \resizebox{\textwidth}{!}{
    \begin{tabular}{lllllllllll}
\toprule
 & {\bf Group 1} & {\bf Group 2} & {\bf Group 3} & {\bf Group 4} & {\bf Group 5} & {\bf Group 6} & {\bf Group 7} & {\bf Group 8} & {\bf Group 9} & {\bf Group 10} \\
\midrule
{\bf All} & 1.000 & 0.864 & 0.939 & 0.866 & 0.875 & 0.880 & 1.000 & 0.600 & 0.996 & 0.903 \\
{\bf BERT} & 1.000 & 0.922 & 0.925 & 0.832 & 0.959 & 0.799 & 0.967 & 0.585 & 0.996 & 0.827 \\
{\bf openSMILE} & 1.000 & 0.922 & 0.900 & 0.832 & 0.880 & 0.839 & 1.000 & 0.947 & 0.996 & 0.827 \\
{\bf CPS} & 1.000 & 0.922 & 0.900 & 0.832 & 0.880 & 0.815 & 0.000 & 0.947 & 0.996 & 0.827 \\
{\bf Action} & 1.000 & 0.922 & 0.900 & 0.832 & 0.880 & 0.873 & 0.571 & 0.947 & 0.996 & 0.827 \\
{\bf GAMR} & 1.000 & 0.922 & 0.900 & 0.832 & 0.880 & 0.731 & 0.658 & 0.947 & 0.996 & 0.827 \\
\bottomrule
	\end{tabular}}
\vspace*{-2mm}
	\caption{\label{tab:avg_sdc_f_or_e-compare-cga}Average DSC per group over FBank $\cup$ EBank, comparing multimodal features and each individual modality.  Propositions are extracted using the CGA method.}
\vspace*{-2mm}
\end{table*}

\begin{figure*}[ht!]
    \centering
    \includegraphics[width=.48\textwidth]{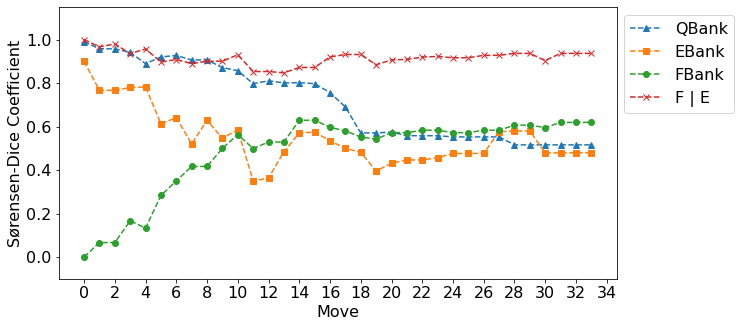}
    \includegraphics[width=.48\textwidth]{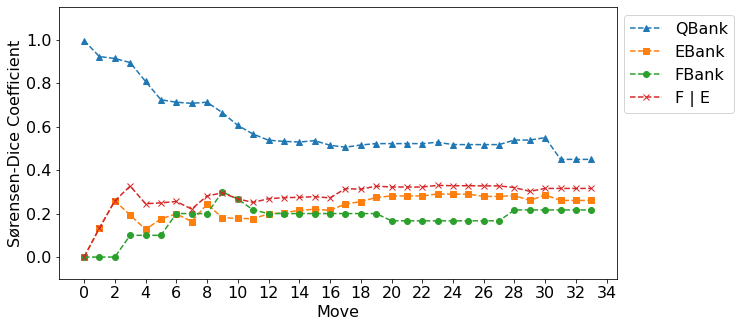}
\vspace*{-2mm}
    \caption{DSC for each bank aggregated across groups, plotted vs. utterance, using all modalities in the move classifier. [L]: propositional extraction performed using the multimodal CGA method. [R]: propositional extraction performed using the language-only Dense Paraphrase (DP) method.}
    \label{fig:all-modalities-cga-vs-dp}
\vspace*{-2mm}
\end{figure*}

Averaged across all groups, the move classification model achieves a weighted F1 of 0.61.  Most misclassifications are confusions of {\it STATEMENT}s and {\it ACCEPT}s, which affect the level of evidence assigned to extracted propositions but not the propositions themselves.

\textbf{Table~\ref{tab:avg_sdc-all_vs_bert-cga}} shows average DSC per group, for each bank, with propositional extraction using the Common Ground Annotation (CGA) method (Sec.~\ref{ssec:prop}).  We also assess the union of the fact bank and evidence bank, to assess how different modalities contribute to the extraction of propositional content and elevation to either status. We compare the performance using all modalities to using language features only, in the form of BERT vectors.

We find that in most cases, our common ground tracker has trouble not with retrieving the right propositions with the multimodal CGA method, but with assigning the right level of evidence.  This is seen in the values for the union of FBank and EBank, which remain high across all groups, even when the S\o rensen-Dice coefficients of the individual FBank or EBank are comparatively lower. This also tracks the misclassifications made by the move classifier, as an {\it ACCEPT} will elevate a proposition to a fact, while a {\it STATEMENT} will keep it in evidence without removing the corresponding QUD from QBank.

Incorprating multiple modalities into the move classifier model usually helps assign propositions to the correct level of common ground and maintain greater overlap in the retrieved QUDs relative to ground truth. However, there is great variety across groups.  For example, Group 2 performs better using only language, while in Group 9 non-linguistic features do not change the result.  These differences can be attributed to how different groups use different modes of communication to complete the task (see Sec.~\ref{sec:disc}).

\textbf{Table~\ref{tab:avg_sdc_f_or_e-compare-cga}} shows average DSC per group over FBank $\cup$ EBank, comparing each individual modality vs. all modalities.

We see here that often, each individual modality performs similarly or identically, but in 4 out of 10 groups, using all multimodal features results in the highest performance. However, in other groups, multimodal features may make no difference, or some other individual feature type is the strongest predictor of performance.  Compare Group 1 and Group 9, where all modalities perform identically, with Groups 6 and 7, where they all perform differently but multimodal features perform the highest. This supports the previous observation: different groups may adopt radically different modal combinations to communicate equivalent information.

\textbf{Fig.~\ref{fig:all-modalities-cga-vs-dp}} shows the progression of DSC over time, aggregated across all groups, using all modalities in the move classifier, but comparing and contrasting the multimodal Common Ground Annotation (CGA) and language model-based Dense Paraphrase (DP) methods for proposition extraction.  Including multimodal information improves the retrieval of the correct propositions independent of the level of evidence or factuality assigned to them---as shown by the consistently high DSC of FBank $\cup$ EBank in the left plot.

\vspace*{-2mm}
\section{Discussion}
\label{sec:disc}
\vspace*{-2mm}

Some specific examples show how the language model-based DP method struggles to extract propositions from complex utterances. Table~\ref{tab:dp-props} shows how vector comparison over only linguistic information tends to struggle with propositions involving multiple objects.  Certain groups, like Group 1, tended to speak full propositions aloud, while others, like Group 10, mixed modalities (``ten and ten'' is accompanied by gesture and action, which are accounted for directly by Common Ground Annotations but not Dense Paraphrases).

\begin{table}[h!]
    \resizebox{.49\textwidth}{!}{
    \begin{tabular}{lll}
\toprule
\textbf{Group} & \textbf{Utterance (DP)} & \textbf{Proposition (Correct?)} \\
\midrule
\multirow{1}{*}{1} & red block's ten so then & \multirow{1}{*}{red = 10 (\cmark)} \\
\multirow{2}{*}{1} & yeah ok so now we know that blue & \multirow{2}{*}{blue = 10 (\cmark)} \\
& block is also ten & \\
\multirow{2}{*}{5} & so red block, blue block are both ten & red = 20 and green = 40 \\
& in theory ten ten twenty & and purple = 10 (\xmark) \\
\multirow{2}{*}{5} & so the green we think is twenty ok so & \multirow{2}{*}{green = 20 (\cmark)} \\
& let's see we can use our hands as well & \\
\multirow{2}{*}{10} & i guess green block is like twenty and & red = 50 and green = 20 \\
& red block, blue block is like ten and ten & and purple = 10 (\xmark) \\
\bottomrule
	\end{tabular}}
\vspace*{-2mm}
	\caption{\label{tab:dp-props}Utterances and propositions retrieved using DP method.}
\vspace*{-2mm}
\end{table}

This reflects Table~\ref{tab:avg_sdc_f_or_e-compare-cga} where, even using the multimodal CGA extraction method, Group 1 achieves perfect overlap of FBank $\cup$ EBank with ground truth using just language, while the model has to combine modalities to reach its best performance for Group 10.  That Group 1 performance over FBank $\cup$ EBank is also perfect using each individual modality alone suggests that their utterances are strongly aligned with their non-verbal behavior.  Meanwhile, Group 6 stands out as a particular case where each individual modality is contributing something distinct.

Misclassifications of {\it STATEMENT}s as {\it ACCEPT}s, or vice versa, may elevate the utterances of certain participants to fact status, or leave elements in the Questions Under Discussion when they have already been resolved. This could also lead to a certain participant having more apparent influence over the dialogue. One participant's beliefs may update the common ground of the group, and leave other participants' beliefs unconsidered. Table~\ref{tab:mislabels} shows some examples from Group 10, and demonstrates how affirmative language like ``yeah'' may be indicative of {\it ACCEPT}s elsewhere in the training data, while ``okay'' or restatements of propositional content are typically indicative of {\it STATEMENT}s even when in context they indicate acceptance of a previously-stated proposition.   

\begin{table}[h!]
    \resizebox{.49\textwidth}{!}{
    \begin{tabular}{llll}
\toprule
\textbf{Timestamp} & \textbf{Utterance} & \textbf{Label} & \textbf{Prediction} \\
\midrule
\multirow{1}{*}{117.46-118.87} & yeah they're together. & \multirow{1}{*}{{\it STATEMENT}} & \multirow{1}{*}{{\it ACCEPT}} \\
\multirow{1}{*}{217.89-219.78} & thirty one thirty two so thirty & \multirow{1}{*}{{\it ACCEPT}} & \multirow{1}{*}{{\it STATEMENT}} \\
\multirow{1}{*}{218.23-219.00} & so okay & \multirow{1}{*}{{\it ACCEPT}} & \multirow{1}{*}{{\it STATEMENT}} \\
\bottomrule
	\end{tabular}}
\vspace*{-2mm}
	\caption{\label{tab:mislabels}Sample of utterances from Group 10 misclassified by move classifier.}
\vspace*{-2mm}
\end{table}

\vspace*{-2mm}
\section{Conclusion and Future Work}
\label{sec:conc}
\vspace*{-2mm}

In this paper we have presented a challenging novel task: {\it multimodal common ground tracking}, and a novel benchmark over the challenging Weights Task Dataset.  We presented a formal model of common ground over a shared task and augmented the WTD with additional gesture, action, and common ground annotations. We performed a set of experiments to evaluate the contributions of different modalities toward modeling the cognitive states of the group, extracting the propositions expressed, and building common ground structures as the group proceeds through the task.  Our model will be particularly useful for AI systems deployed in environments such as classrooms, where they can track the collective knowledge of a group and facilitate productive collaborations.

Certain modalities may be more prone to misclassifications based on the speaker.  For instance, future work could examine how prosidic features could be used to detect power dynamics that may bias the construction of common ground toward certain people or assertions.  Giving the common ground model additional separate banks for each speaker would allow an agent to facilitate knowledge sharing and collaboration if it seems like a subgroup has arrived at a belief not shared by the whole group.  In a task-based environment, the agent could use the model of common ground to make task-relevant inferences itself, such as the algebraic relationship between the block weights here, allowing it to learn from watching and interacting with the group.  Finally, because there is a one-to-many mapping between propositions and potential ways to phrase or express them in utterances, the dense paraphrase method for propositional extraction could benefit from a cross-encoder approach, as used in coreference research.

\section*{Limitations}
Although our work addresses a novel and challenging problem, scaling the pipeline to other use cases confronts some (surmountable) limitations. For a given task, the relevant propositions that may populate the common ground need to be determined and enumerated. The number of propositions scales naturally to increased cardinality of items, attributes, and relations within a similar domain (e.g., by computing the Cartesian product of items, attributes, and binary relations, and subsequently the powerset of atomic propositions to account for conjunctions like $red = 10 \wedge blue = 10$). Therefore the complexity of proposition construction is subject to the complexity of the task and number of task items. Enumerating the closure rules is straightforward once the propositions are determined. The move classifier itself should require no changes unless there is a change in input modalities. Imbalance within the data categories presents a further challenge that needs to be addressed. In this paper we used data augmentation approaches like SMOTE, but precise handling would need to be determined on a task-specific basis.

\section*{Acknowledgements}

This work was partially supported by the National Science Foundation under award DRL 2019805 to Colorado State University and Brandeis University, and awards IIS 2326985 and IIS 2033932 to Brandeis University. The views expressed are those of the authors and do not reflect the official policy or position of the U.S. Government. All errors and mistakes are, of course, the responsibilities of the authors. Special thanks to Jade Collins, August Garibay, and Carlos Mabrey for extensive data annotation.

%Place all acknowledgments (including those concerning research grants and funding) in a separate section at the end of the paper.

% \input{charts/avg_sdc-all_modalities-cga}

% \input{charts/avg_sdc-all_modalities-dp}

% \input{charts/avg_sdc-bert-cga}

% \input{charts/avg_sdc-bert-dp}

% \input{charts/avg_sdc-opensmile-cga}

% \input{charts/avg_sdc-opensmile-dp}

% \input{charts/avg_sdc-cps-cga}

% \input{charts/avg_sdc-cps-dp}

% \input{charts/avg_sdc-action-cga}

% \input{charts/avg_sdc-action-dp}

% \input{charts/avg_sdc-gamr-cga}

% \input{charts/avg_sdc-gamr-dp}

%\newpage
\section*{Bibliographical References}\label{reference}

\bibliographystyle{lrec_natbib}
\bibliography{coling-cgt}

% \section{Language Resource References}
% \label{lr:ref}

% \bibliographystylelanguageresource{lrec_natbib}
% \bibliographylanguageresource{languageresource}

\appendix

\section{Group-wise Move Classifier Performance}
Table~\ref{tab:metrics} shows the performance of the 10 classifiers with each being trained on 9 different groups, and evaluated on the remaining 10th.

\begin{table*}[h!]
\centering
\small
\begin{adjustbox}{width=\textwidth}
\begin{tabular}{l*{10}{p{0.9cm}}}
\toprule
 & {\bf Group 1} & {\bf Group 2} & {\bf Group 3} & {\bf Group 4} & {\bf Group 5} & {\bf Group 6} & {\bf Group 7} & {\bf Group 8} & {\bf Group 9} & {\bf Group 10} \\
\midrule
\bf Accuracy & 0.625 & 0.720 & 0.750 & 0.500 & 0.735 & 0.690 & 0.429 & 0.556 & 0.833 & 0.438 \\ 
\cdashline{1-11}
\bf Micro F1 & 0.625 & 0.720 & 0.750 & 0.500 & 0.735 & 0.690 & 0.429 & 0.556 & 0.833 & 0.438 \\ 
\bf Macro F1 & 0.564 & 0.360 & 0.621 & 0.235 & 0.512 & 0.375 & 0.271 & 0.389 & 0.652 & 0.203 \\ 
\bf Weighted F1 & 0.605 & 0.680 & 0.768 & 0.412 & 0.706 & 0.737 & 0.453 & 0.509 & 0.791 & 0.342 \\ 
\cdashline{1-11}
\bf Micro Precision & 0.625 & 0.720 & 0.750 & 0.500 & 0.735 & 0.690 & 0.429 & 0.556 & 0.833 & 0.438 \\ 
\bf Macro Precision & 0.583 & 0.369 & 0.611 & 0.200 & 0.525 & 0.382 & 0.288 & 0.355 & 0.912 & 0.167 \\ 
\bf Weighted Precision & 0.604 & 0.654 & 0.796 & 0.350 & 0.687 & 0.837 & 0.526 & 0.473 & 0.863 & 0.281 \\ 
\cdashline{1-11}
\bf Micro Recall & 0.625 & 0.720 & 0.750 & 0.500 & 0.735 & 0.690 & 0.429 & 0.556 & 0.833 & 0.438 \\ 
\bf Macro Recall & 0.567 & 0.365 & 0.650 & 0.286 & 0.516 & 0.462 & 0.298 & 0.433 & 0.625 & 0.259 \\ 
\bf Weighted Recall & 0.625 & 0.720 & 0.750 & 0.500 & 0.735 & 0.690 & 0.429 & 0.556 & 0.833 & 0.438 \\ 
\cdashline{1-11}
\bf AUROC & 0.500 & 0.539 & 0.500 & 0.563 & 0.500 & 0.501 & 0.476 & 0.669 & 0.500 & 0.531 \\ 
\bottomrule
\end{tabular}
\end{adjustbox}
\vspace*{-2mm}
\caption{Group-wise performance of the move classifier using hold-one-group-out evaulation method.}
\vspace*{-2mm}
\label{tab:metrics}
\end{table*}

\section{Annotation Procedures and IAA}
\label{app:anno}

\begin{figure*}
    \centering
    \includegraphics[width=0.9\textwidth]{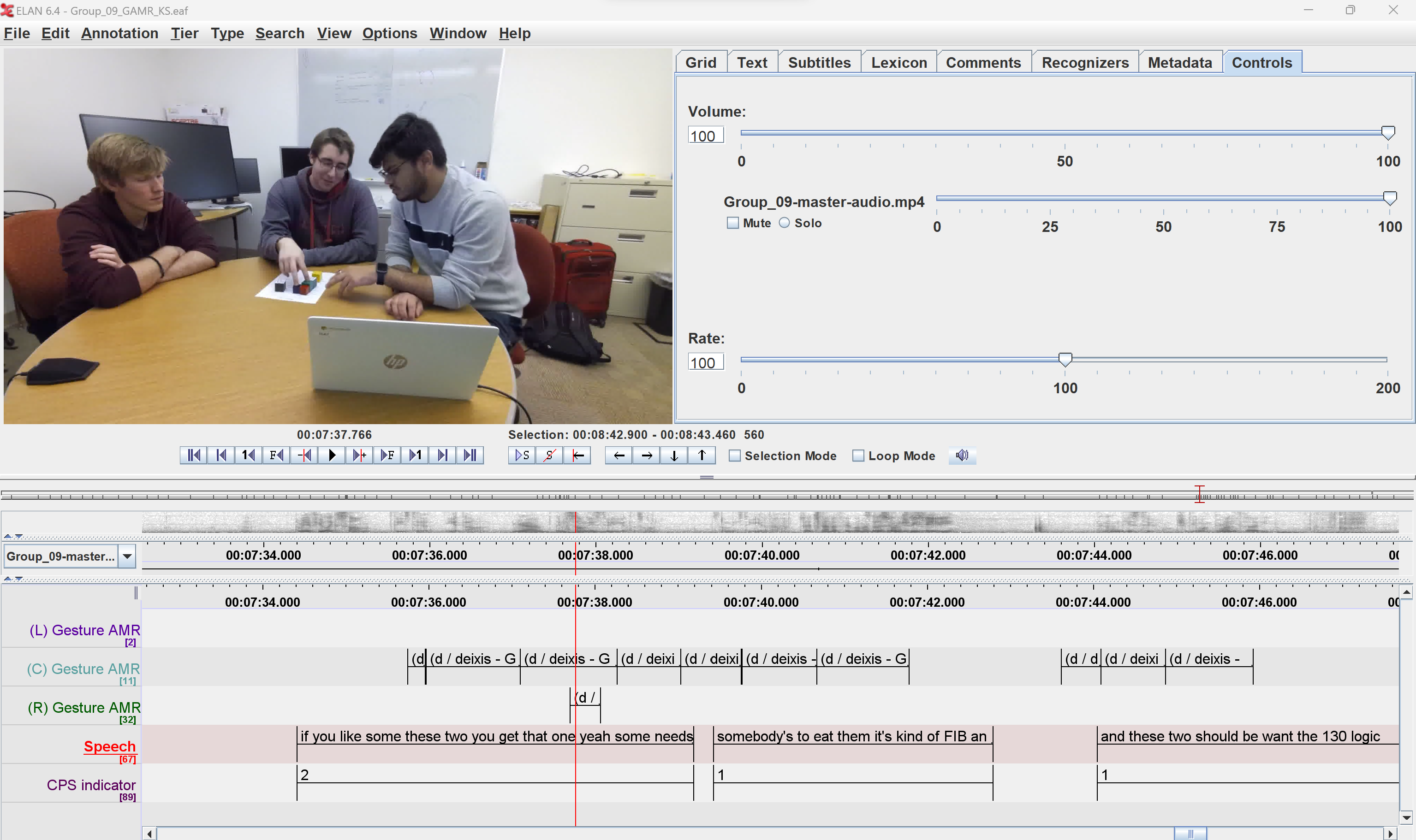}
% \vspace*{-2mm}
    \caption{\label{fig:Elan}Still of annotation procedure using ELAN.}
\vspace*{-2mm}
\end{figure*}

ELAN~\cite{brugman2004annotating} was the tool used for most annotation, supplemented by collating annotations in spreadsheets. As we can see in Fig.~\ref{fig:Elan}, this tool allows annotators to visualize the data at any point in the videos, and also see all other annotated modalities. The data is then featurized and used as input for the move classifier.

Tables~\ref{tab:cga-iaa}--\ref{tab:gamr-iaa} show inter-annotator agreement (IAA) metrics for the Common Ground, action, and GAMR annotation per group. Because gestures are individualized, GAMR annotations are also broken down by participant. Means are also provided.

\begin{table}[htb]
\begin{center}
\begin{tabular}{lll}
      \toprule
      \bf Group & \bf F1 score & \bf Cohen's $\kappa$ \\
      \midrule
      \bf 1 & 0.520 & 0.359 \\
      
      \bf 2 & 0.454 & 0.295 \\
      
      \bf 3 & 0.492 & 0.356 \\
      
      \bf 4 & 0.411 & 0.267 \\
      
      \bf 5 & 0.471 & 0.503 \\
      
      \bf 6 & 0.639 & 0.603 \\
      
      \bf 7 & 0.678 & 0.572 \\
      
      \bf 8 & 0.522 & 0.712 \\
      
      \bf 9 & 0.575 & 0.564 \\
      
      \bf 10 & 0.645 & 0.772 \\
      
      \bf mean & 0.541 & 0.500 \\
      \bottomrule

\end{tabular}
\caption{\label{tab:cga-iaa}IAA on Common Ground Annotations.}
 \end{center}
\end{table}

\begin{table}[htb]
\begin{center}
\begin{tabular}{lll}
      \toprule
           \bf Group & \bf F1 score & \bf Cohen's $\kappa$ \\
      \midrule
      
      \bf1 & 0.557 & 0.464 \\
      
      \bf2 & 0.651 & 0.666 \\
      
     \bf3 & 0.750 & 0.688 \\
     
     \bf4 & 0.719 & 0.654 \\
     
     \bf5 & 0.804 & 0.689 \\
     
   \bf  6 & 0.737 & 0.798 \\
     
   \bf  7 & 0.761 & 0.660 \\
     
   \bf  8 & 0.583 & 0.466 \\
     
  \bf   9 & 0.519 & 0.432 \\
     
  \bf  10 & 0.629 & 0.458 \\
     
  \bf   mean & 0.671 & 0.597 \\
     \bottomrule

\end{tabular}
\vspace*{-2mm}
\caption{\label{tab:action-iaa}IAA on action annotations.}
\vspace*{-2mm}
 \end{center}
\end{table}

\begin{table}[!ht]
\begin{center}
\resizebox{.8\columnwidth}{!}{
\begin{tabular}{lllll}
      \toprule
      \bf Group & \bf Participant &\bf F1& \bf Precision & \bf Recall\\
      \midrule
      
      1 & 1 & 0.921 & 0.953 & 0.890 \\
      
      1 & 2 & 0.943 & 0.917 & 0.971 \\
      
      1&3 & 0.899 & 0.912 & 0.886\\
      
      1&$\mu$ & 0.921 & 0.927 & 0.915 \\
      \cdashline{1-5}
      
     2 & 1 & 0.846 & 0.798 & 0.902 \\
      
      2 & 2 & 0.947 & 0.938 & 0.957 \\

      2&3 & 0.895 & 0.850 & 0.944\\
     
     2&$\mu$ & 0.896 & 0.862 & 0.934 \\
      \cdashline{1-5}
        
      3 & 1 & 0.686 & 0.720 & 0.656 \\
      
      3 & 2 & 0.809 & 0.796 & 0.824\\
      
      3&3 & 0.793 & 0.775 & 0.811\\
           
    3&$\mu$ & 0.763 &0.763 & 0.763 \\
      \cdashline{1-5}
        
        4 & 1 & 0.791 & 0.837 & 0.750 \\
      
      4 & 2 & 0.658 & 0.807 & 0.556\\
      
      4&3 & 0.817 & 0.779 & 0.859\\
           
    4&$\mu$ & 0.755 &0.808 & 0.722 \\
      \cdashline{1-5}
          
        5 & 1 &  0.824 & 0.836 & 0.813 \\
      
      5 & 2 & 0.693 & 0.642 & 0.754\\
      
      5&3 & 0.835 & 0.853 & 0.817\\
           
    5&$\mu$ & 0.784 &0.777 & 0.795 \\
      \cdashline{1-5}

     6 & 1 &  0.697 & 0.704 & 0.691 \\
      
      6 & 2 & 0.628 & 0.667 & 0.594\\
      
      6&3 & 0.480 & 0.462 & 0.500\\
           
    6&$\mu$ & 0.602 &0.611 & 0.595\\
      \cdashline{1-5}
          
    7 & 1 & 0.865 & 0.874 & 0.857 \\
      
      7 & 2 & 0.736 &0.724 & 0.748\\
      
      7&3 & 0.667& 0.662& 0.671\\
           
    7&$\mu$ & 0.756 &0.753 & 0.759\\
      \cdashline{1-5}
         
    8 & 1 & 0.782 & 0.725 & 0.850 \\
      
      8 & 2 & 0.745 &0.710 & 0.784\\
      
      8&3 & 0.907& 0.925& 0.891\\
           
    8&$\mu$ & 0.812 &0.787 & 0.841\\
      \cdashline{1-5}
          
      9 & 1 & 0.846 & 0.846 & 0.846 \\
      
      9 & 2 & 0.600 &0.525 & 0.700\\
      
      9&3 & 0.800& 0.818& 0.783\\
           
    9&$\mu$ & 0.749 &0.730 & 0.776\\
      \cdashline{1-5}
          
      10 & 1 & 0.386 &	0.810	& 0.254 \\
      
      10 & 2 & 0.487 &	0.631	& 0.396\\
      
      10&3 & 0.584	&0.444	&0.851\\
           
    10&$\mu$ & 0.749 &0.730 & 0.776\\
      \cdashline{1-5}
         
 $\mu$ & 1 & 0.765 &	0.754	& 0.806 \\
      
      $\mu$ & 2 & 0.738 &	0.712	& 0.752\\
      
      $\mu$ &3 & 0.768	& 0.789	&0.761\\
           
    $\mu$ &$\mu$ & 0.752 & 0.752 & 0.773\\
        \bottomrule

\end{tabular}}
\vspace*{-2mm}
\caption{\label{tab:gamr-iaa}IAA on Gesture-AMR (GAMR) annotation.}
\vspace*{-2mm}
 \end{center}
\end{table}
% \newpage
% \vspace{10mm}

\end{document}